\documentclass[a4paper]{article}

\usepackage{INTERSPEECH2021}
\usepackage{tipa}
\usepackage{color,soul}
\usepackage{xspace}
\usepackage{hyperref}
\usepackage{hyphenat}
\usepackage{subfigure}

\title{Experiments of ASR-based mispronunciation detection for children and adult English learners}
\name{Nina Hosseini-Kivanani* \thanks{*Erasmus Mundus funded joint degree in Language and Communication
Technologies.}$^1$ $^,$ $^2$, Roberto Gretter$^3$, Marco	Matassoni$^3$, Giuseppe Daniele	Falavigna$^3$}

\address{
  $^1$ University of Trento \\
  $^2$ University of Groningen \\
  $^3$ Fondazione Bruno Kessler (FBK)}
\email{nina.hossini@gmail.com, \{gretter, matasso, falavi\}@fbk.eu} 

\begin{document}

\maketitle
\begin{abstract}
Pronunciation is one of the fundamentals of language learning, and it is considered a primary factor of spoken language when it comes to an understanding and being understood by others. 
The persistent presence of high error rates in speech recognition domains resulting from mispronunciations motivates us to find alternative techniques for handling mispronunciations. In this study, we develop a mispronunciation assessment system that checks the pronunciation of non-native English speakers, identifies the commonly mispronounced phonemes of Italian learners of English, and presents an evaluation of the non-native pronunciation observed in phonetically annotated speech corpora. 
In this work, to detect mispronunciations, we used a phone-based ASR implemented using Kaldi.
We used two non-native English labeled corpora;
\begin{enumerate*}[label=(\roman*), itemjoin={{, }}, itemjoin*={{, and }}]
    \item a corpus of Italian adults contains 5,867 utterances from 46 speakers
    \item a corpus of Italian children consists of 5,268 utterances from 78 children. 
\end{enumerate*} Our results show that the selected error model can discriminate correct sounds from incorrect sounds in both native and nonnative speech, and therefore can be used to detect pronunciation errors in non-native speech. The phone error rates show improvement in using the error language model. The ASR system shows better accuracy after applying the error model on our selected corpora. 
\end{abstract}

\section{Introduction}
The number of people who are learning a second language (L2) worldwide is increasing. Consequently, the need to evaluate and grade their pronunciation is becoming an important topic.
The most challenging part of learning a new language is the pronunciation part due to the fact that it is challenging to imitate sounds that are different to those of the native language's phoneme inventory~\cite{Huensch2019,McCrocklin2016, Peabody2011}.~\cite{Fant1973} defines mispronunciation as surface pronunciation forms differing from canonical pronunciation forms. \textbf{Phoneme level mispronunciation} refers to the interference of a second language learner's native language during speech production, where foreign sounds are produced similar to a phoneme in their native language. Most of the classroom's pronunciation activities rely on the teacher to monitor, evaluate, and provide feedback on student pronunciation. This traditional technique does not seem adequate to correct student pronunciation, and it tends to be costly and time-consuming.

Given these constraints, the growing tendency to assess non-native languages leads to increased interest in automatic proficiency assessment of speech and boosts Computer Assisted Language Learning (CALL) tools in the field of language teaching for L2 learners~\cite{Witt2012a, Neri2003, kawai1998call}. CALL tools are designed to recognize words or sentences uttered by L2 learners by using an ASR system. The CAPT system is one of CALL's essential tools designed for automatically evaluating and detecting the learner's pronunciation errors. In the CAPT system, pronunciation evaluation can happen at two levels~\cite{Peabody2011}:
\begin{enumerate*}[label=(\roman*), itemjoin={{, }}, itemjoin*={{, and }}]
    \item detecting specific pronunciation errors
    \item an overall assessment of a speaker's proficiency (i.e., goodness of pronunciation (GoP)~\cite{Zechner2009, downey2008evaluation, Witt2000}).
\end{enumerate*}

This study seeks to examine the use of ASR to empower learners to practice and improve their pronunciation on their own. Using a well-designed ASR system allows students to work autonomously while also offering flexibility to improve their pronunciation. The persistent presence of high error rates in speech recognition domains resulting from mispronunciations motivates us to find alternative techniques for handling mispronunciations. We use two corpora (for more information, see section~\ref{label:data}) of both children and adult Italian speakers. These corpora contain both spontaneous and read-aloud tasks.

The innovation of our system is to use error rules in our language model (error language model) based on the most common errors that were seen in L2 learners' speech, in order to improve our detection system (see section~\ref{sect:method}). Error detection of mispronounced input is similar to the process in which an annotator manually indicates pronunciation errors. Mainly, the topic of interest in this study is the mispronunciation assessment at the phone level. Ideally, our system should be able to detect errors the same as human annotators do. 

\section{Related Work}
\label{sect:related_work}
Segmental features have been a subject of phonetics from the 1950s up to now~\cite{Ryu2017}. The most common segmental features for investigating the pronunciation of L2 learners on automatic speech assessment are consonant features: stop closure duration, aspiration, and vowel features: vowel duration. In one of the first studies on the assessment of pronunciation at a phone level,~\cite{@phdthesiswitt1999use2019} found that the scoring accuracy for the assessment of errors was 80-92$\%$ for non-native English speakers (73 speakers).

Early research by~\cite{FrancoHoracioandNeumeyerLeonardoandKimYoonandRonen1997} used two kinds of acoustic models to conduct automatic mispronunciation detection: \begin{enumerate*}[label=(\roman*), itemjoin={{, }}, itemjoin*={{, and }}]
    \item a model trained on native-speaker pronunciation
    \item a model trained on non-native speech.
\end{enumerate*}
They used an acoustic model to calculate the log probability for each predicted phone in both models. They then calculated the difference between these two probability scores and used it as a metric for rating the pronunciation's quality. The result showed that the log-likelihood ratioLLR had a better performance than the log-posteriors method. According to~\cite{KimYoonandFrancoHoracioandNeumeyer1997}, posterior probabilities and log-likelihood scores were the methods that were most correlated with word and phone level human assessments of pronunciation.

As an example of the earliest work in this field, we can refer to the FLUENCY system~\cite{Eskenazi1999} to detect pronunciation problems at both the phonetic and the prosodic levels (i.e., segmental and suprasegmental levels). They used the SPHINX-II speech recognizer to evaluate and detect phone errors and prosodic information (namely prosodic problems) of non-native speakers of French, German, Hebrew, Hindi, Italian, Mandarin, Portuguese, Russian, and Spanish, who learned English as their second language. It was reported that using ASR technology while learning a foreign language can reduce embarrassment and enhance learning for learners.

~\cite{Tao2016} investigated three ASR systems for nativeness evaluation in their study: a GMM-HMM system, a DNN-HMM system, and a GMM-HMM system using DNN for feature extraction. The feature sets were categorized into fluency, rhythm, pronunciation, grammar, and vocabulary for segmental (at the consonant and the vowel level) and suprasegmental measurements of non-native speech in automatic non-native speech assessment. We will base our work on these features in our ASR system. 

The above studies are a small set of examples that are related to segmental features. We focus on segmental features, in which we observe the specific phoneme-level errors made by second language learners. For learners, the most challenging part of learning a new language is the pronunciation part, due to the fact that it is challenging to imitate sounds that are different to those of the native language's phoneme inventory~\cite{Peabody2011}. Using a well-designed ASR system allows students to work autonomously while also offering flexibility that can lead to the improvement of their pronunciation. 

\section{Methods}
\label{sect:method}
In the following sections, we describe the methods (language model and acoustic model) used to identify pronunciation errors made by Italian learners of English, to develop an ASR system to improve their L2 pronunciation. 

\subsection{Speech file transcriptions: Ideal, Manual, \& ASR}

For this study, the three transcriptions at the phone level were considered: Ideal, Manual, and ASR output. All the transcriptions are automatically time-aligned to the related speech signals: 
\begin {itemize}
\setlength\itemsep{0.15em}
\item
\textbf{Ideal (IDE)} (reference) refers to the canonical word's pronunciation. 
\item 
\textbf{Manual (MAN)} (reference) is the hand-corrected version of the ideal data at the phoneme level. Trained annotators corrected files manually using Praat tool~\cite{Boersma2016}, noting substitution, insertion, and deletion at the phoneme level.
\item
\textbf{ASR output} is the results of ASR processing given a set of phones as input (see section~\ref{sec:LM}).
\end{itemize}

Given these three representations (i.e., IDE vs.\ MAN, IDE \& ASR, and MAN \& ASR), the comparison between specific pairs of phone sequences can provide us the following information: 

\textbf{IDE vs.\ MAN output}: this gives us a true map of the errors made by Italian speakers when trying to speak English. The comparison of IDE \& MAN could be used to build some models of the errors made by L2 speakers.; \textbf{IDE vs.\ ASR output}: this gives us a feeling of what an automatic system can detect for a new utterance, where no  manual annotation is available; \textbf{MAN vs.\ ASR output}: this tells us how well an automatic system detects errors.
The mispronunciation at the segmental level (i.e., phonetic) was categorized into three kinds at the phone level: substitution (i.e., a phoneme is replaced with another), insertion (i.e., an extra phoneme is inserted), and deletion (a certain phoneme is deleted)~\cite{Peabody2011}.

\subsection{Proposed approach}
\subsubsection{Language model}\label{sec:LM}
The basic idea of language models is to provide a probability of a sentence or sequence of words; these probabilities are combined with the acoustic likelihood of the sequence and generate the resulting hypothesis~\cite{Jurafsky2019}.
In this work \textbf{phone n-grams} are used as a language model to allow the generation of phone sequences not constrained by only the words appearing in the lexicon; $n$-grams are trained on the phonetic transcription of the audio data, with $n$ ranging from 1 to 5.

Apart from the n-gram model, another novel approach is the application of an \textbf{error language model} based on the most frequent errors (hereafter error model). For the error model, we used lexical information by providing the ASR with the canonical pronunciations of the (known) words to be recognized and other pronunciations obtained by applying phonetic rules. These rules were defined manually by looking at the most common errors resulting from the 5-gram phone model (Table~\ref{tab:error_rule}). 

\subsubsection{Acoustic Model}
The acoustic model is learned from a set of audio recordings and their corresponding transcripts.
We trained hybrid GMM-DNN models on English and Italian speech data from the child and adult Italian speakers. Based on the common Kaldi recipe \footnote{Kaldi: \url{http://kaldi-asr.org/doc/}}, the selected acoustic models' features are: \begin{enumerate*}[label=(\roman*), itemjoin={{, }}, itemjoin*={{, and }}]
    \item initial GMM models trained on MFCC acoustic features with LDA transform and speaker adaptive training \cite{MiaoYajieandZhangHaoandMetze2015}
    \item use of i-vectors \cite{MadikeriSrikanthandDeySubhadeepandMotlicekPetrandFerras2016} of size 100, stacked to the MFCCs 
    \item TDNNs trained using LF-MMI \cite{povey2016purely}
\end{enumerate*}. 

\section{Data}
\label{label:data}
The description of each corpus will be given below;
\begin{enumerate*}[label=(\roman*), itemjoin={{, }}, itemjoin*={{, and }}]
    \item a corpus of Italian children (ChildEn)~\footnote{This corpus was designed and collected by ITC-irst: ~\url{http://www.itc.it}}~\cite{Batliner2005}
    \item  corpus of Italian adults (ISLE)~\cite{Menzel2000a}
\end{enumerate*}, who are learning English as their second language.
Detecting mispronunciations in ASR requires corpora with labeling at the phonetic level: we selected these two corpora because they were manually labeled in terms of pronunciation quality by humans, and both were manually transcribed at phone-level.

\textbf{ChildEn} consists of 5,268 utterances, with an overall duration of 3h:28m:26s from 78 children at about ten years of age. The selected students had been studying English at school for 3 or 4 years. 

\textbf{ISLE} is a non-native speech dataset that contains utterances recorded by intermediate-level adult German and Italian learners of English. The audio files are 17h:54m:44s in total. The Italian section contains 23 Italian speakers.

\subsection{Phone set}
We created a phone list of each phoneme that includes canonical pronunciation and every possible mispronunciation of English phones by L2 learners; in other words, we have created a phoneme dictionary that contains both English and Italian phones in order to be able to capture all possible mispronunciations.
According to~\cite{Browning2004}, Italian and English share only 40\% of their phones.
Therefore, we expect to see more phonological interference from Italian when L2 learners need to pronounce the phones in English.
Moreover, in Italian, the relationship between spelling and pronunciation is straightforward compared to English. For example, in Italian, the letter 'a' is pronounced /a/, but in English, pronunciation and spelling are not strictly related to each other. For example the letter 'u' is pronounced /u/, /V/ or /\textrevepsilon/ in English\footnote{\url{http://archive.is/zsxA}}.

\subsection{Evaluation}
\begin{figure}[htp!]
\mbox{
\centering
 \begin{subfigure}{
 \includegraphics[width=1.45in, height = 1.9in]{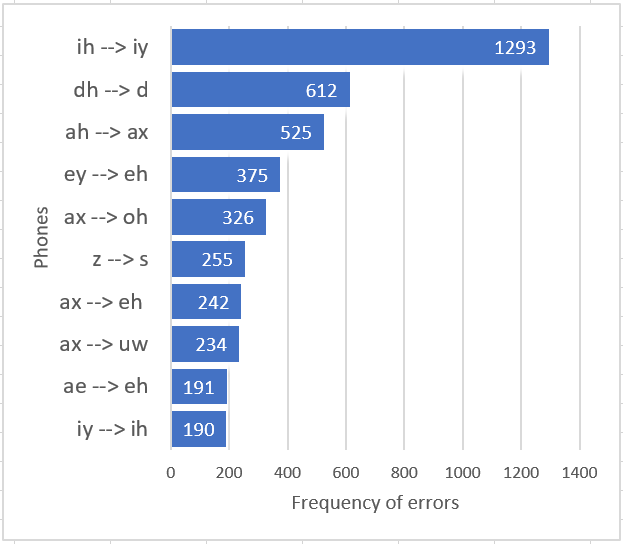}}
 \end{subfigure}\quad
 \begin{subfigure}{
\includegraphics[width=1.45in, height= 1.9in]{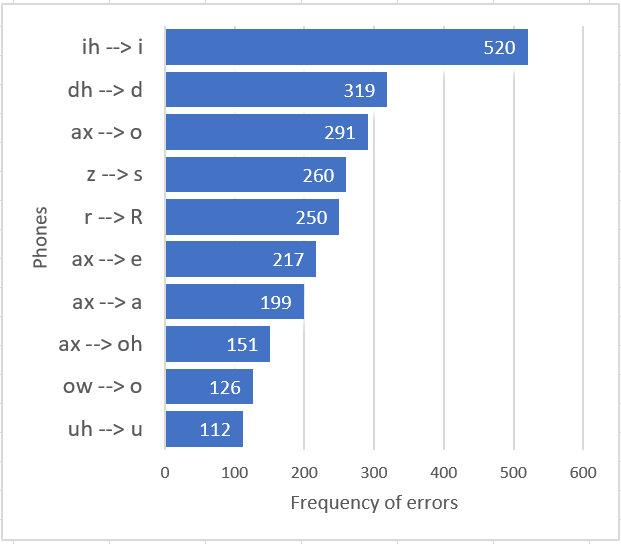}}
 \end{subfigure}\quad}
 \caption{Phoneme errors distributions of ISLE (left graph) \& ChildEn (Right graph).}\label{fig:ISLE_child_top}
\end{figure}

According to Figure~\ref{fig:ISLE_child_top}, the most common mispronunciations for Italian speakers occur when the English target phone is not in the phoneme set of Italian (e.g., English phoneme /ax/). These phones are not contrastive for Italian speakers, leading to the speaker attempting to substitute the phonetically-closest phone from the Italian phoneme set or delete that phone. By ``close", we mean that the acoustic signal has similar formants in both languages or the orthographic representation (spelling) is similar to spelling in Italian.

~\cite{Witt2012a} provided a list of pronunciation errors that can be considered in ASR studies: Phonemic deletion, phonemic insertion, and phonemic substitution. 
For assessing the performance of a phoneme, the phone error rate (PER) will be used. The phone-error rate calculated the log-likelihood of a predicted phoneme given the acoustic signal~\cite{KimYoonandFrancoHoracioandNeumeyer1997, KawaiGohandHirose1998}.
The PER takes into account the errors related to phoneme substitutions (S), phoneme deletions (D), phoneme insertions (I), and P stands for the number of phones. If there are P phones in the reference transcript, and the ASR output has S, D, and I, then multiply by 100:
\begin{equation}\label{eq:per}
\itemsep0em 
\mathrm{PER}=\frac{\mathrm{S}+\mathrm{D}+\mathrm{I}}{\mathrm{P}}*100
\end{equation}
\section{Results}
\label{label:results}
\subsection{Detection output}
In this study, we are concerned with identifying mispronounced phonemes by L2 learners and improving our ASR system; we are interested in the PER and accuracy of our system. 

We used Kaldi to perform speech recognition for phoneme error detection using the dictionary of words and phonemes. The acoustic model was trained with mixed speech data (i.e., children and adults), and the utterances were force-aligned based on the newly adapted word transcription. In our ASR system, we considered the following procedures in the acoustic model;
\begin{enumerate*}[label=(\roman*), itemjoin={{, }}, itemjoin*={{, and }}]
    \item speech recognition based on the phone-level n-gram model 
    \item forced-alignment based on the existing word transcriptions, including the transcriptions modified using the mispronunciation data 
    \item GMM classifier on using the native and the non-native acoustic model.
\end{enumerate*}
The following sections compare the output of phoneme sequences to the gold standard, which leads us to identify pronunciation errors.
\subsubsection{N-gram output}
The ASR was run on 1-gram, 2-gram, 3-gram, 4-gram, and 5-gram of phones. We first trained the algorithm introduced on our data to obtain the baseline values for the top frequent errors to determine the rules for the error model.

According to~\cite{Jurafsky2019}, the common n-gram models, when there is sufficient training data, are trigram, 4-gram, or even 5-gram. We trained all the n-gram models, but we saw that performance was much better with the 5-gram language model. For the rest of the n-grams, we only report the PERs for ASR \& MAN, ASR \& IDE, and IDE \& MAN. The reason is that the 5-gram model captures more context than other models, and thus we chose to report in full detail only the 5-gram results based on the number of substitutions, insertions, and deletions. For this reason, the 5-gram model was used for finding modified transcriptions. The PERs of 5-gram in ISLE and ChildEn are 42\% and 38\% for ASR \& MAN comparison, respectively.  As we have enough data for training, we considered the output of 5-gram models to develop our error model rules. The ASR systems' performance that used n-gram models was low; the PER for each n-gram model was between 33-52\%.

\subsubsection{Error model output}
Thus far, we have tried to improve the PER to estimate the ASR system's performance for mispronunciation recognition of L2 learners of English. The second model used the predefined rules (Table~\ref{tab:error_rule}) from the n-gram model to train native and non-native speech data to check if the output of our ASR system will be improved compared to the n-gram model; we applied the forced-alignment by using the adapted lexicon.

The comparison of our models' performance (n-gram model and error model) was made by calculating PER. We choose the best model based on the PER metric. Interestingly, the new ASR system performs better in terms of PER for ChildEn, and the error rules showed better improvement for the ASR system. The speech recognition phone error rates are typically greater for adults than children. The low improvement in ISLE might be due to more complex and long sentences used in this corpus that lead to error propagation differently. The other possible reason might be due to the recordings' quality (i.e., noise in the background).



\begin{figure}[!ht]
\begin{center}
\label{fig:error_accuracy}
\fbox{\includegraphics[width=2.4in, height = 1.5in]{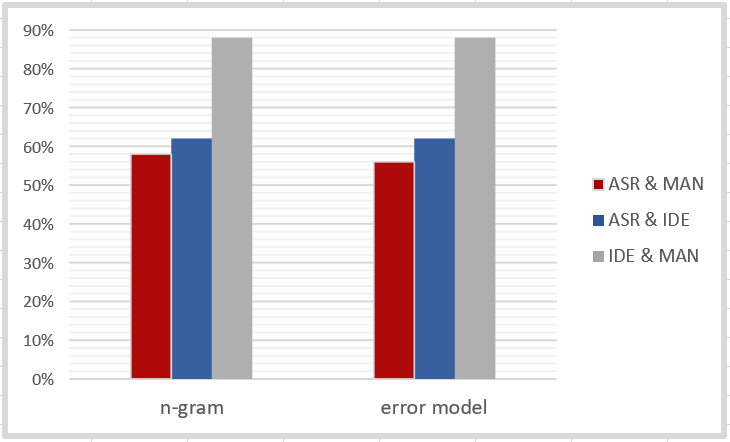}}\\
\vspace{-7pt}
\caption*{\footnotesize\centering{1. ISLE}}
\vspace{3pt}
\fbox{\includegraphics[width=2.4in, height= 1.5in]{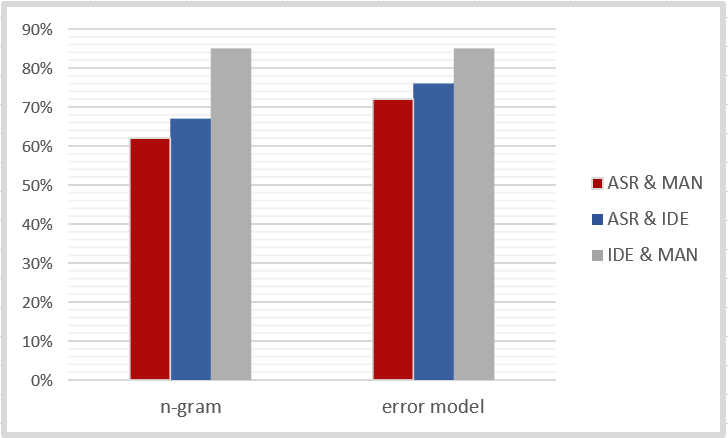}}\\
\vspace{-7pt}
\caption*{\footnotesize{2. ChildEn}}
\caption{Overall detection accuracy for ISLE (top graph). \& ChildIt (bottom graph)\label{fig:accuracy}}
\end{center}
\end{figure}

Overall, the highest Average Accuracy (see Eq.~\ref{eq:per}) to date was obtained using the error model. The accuracy results range from 72\%-76\% for correct error detection of phones in ChildEn (Figure~\ref{fig:accuracy}). 

\section{Discussion}
Focusing on the smaller units would allow students to focus more on specific aspects of their pronunciation. However, the evaluation of smaller units for pronunciation assessment has a higher uncertainty compared to the evaluation of longer units~\cite{Witt2000}. For pronunciation training, especially for L2 learners, automatic pronunciation evaluation can play an important role, for example, by using ASR systems to evaluate the pronunciation of non-native input and quantify how close the speech is to a native-like pronunciation. 

One of the challenges in using non-native speech for automatic recognition is the diversity of allophones, accents, invented words by L2 learners, and longer hesitations. 
We identified high-priority phones (i.e., the phones used for creating the error rules (see Table~\ref{tab:error_rule})
and evaluated our system based on them. Since achieving a perfect native-like pronunciation is an unrealistic goal for adult learners, we focused on the specific mispronounced phones summarized in Table~\ref{tab:error_rule}. As an example:
\begin{itemize}
\itemsep0em 
\small{
\item \textbf{dh} $\rightarrow$ \textbf{dh/d}: ``dh'' can be replaced by itself or ``d''.
\item \textbf{n d} $\rightarrow$ \textbf{n d/}: ``d'' can be deleted if it is in the sequence of ``n d".
\item \textbf{er} $\rightarrow$ \textbf{er r/R/}}: means that our adapted lexicon1 will have three transcriptions for the word “HER” (i.e., ``hh er" (ideal phonetic transcription), ``hh er" \textcolor{red}{R}, \& ``hh er" \textcolor{red}{r}.
\end{itemize}
\begin{table}
\centering
\caption{Error rules: Substitutions, Deletions, \& Insertions}
\label{tab:error_rule}
\resizebox{\columnwidth}{!}{%
\begin{tabular}{|l|l|l|l|l|l|} 
\hline
\multicolumn{2}{|c|}{\textbf{Substitution rules} } & \multicolumn{2}{c|}{\textbf{Deletion rules} } & \multicolumn{2}{c|}{\textbf{Insertion rules} } \\ 
\hline
\multicolumn{1}{|c|}{Original} & \multicolumn{1}{c|}{Sub} & \multicolumn{1}{c|}{Original} & \multicolumn{1}{c|}{Del} & \multicolumn{1}{c|}{Original} & \multicolumn{1}{c|}{Insert} \\ 
\hline
ih   – & ih/i/iy & n d   – & n d/ & er  – & er r/R/ \\ 
\hline
dh   – & dh/d & l d   – & l d/ & aa  – & aa r/R/ \\ 
\hline
ax   – & ax/a/o/oh & th d  – & th d/ & ao  – & ao r/R/ \\ 
\hline
z    – & z/s & s t   – & s t/ & p l – & p ax/o l \\ 
\hline
r    – & r/R & n t   – & n t/ & b l – & b ax/o l \\ 
\hline
uh   – & uh/u & ay k  – & ay k/ & k l – & k ax l \\ 
\hline
 &  & ae k  – & ae k & g l  – & g ax l \\
\hline
\end{tabular}%
}
\end{table}

This model was trained on the phone errors observed in our corpora. We created a set of rules to consider these errors in our corpora and implemented our system based on the adapted lexicon. 
From the output of our ASR system, it is clear that the selected architecture is more capable of detecting mispronounced phones.
In fact, our ASR systems' outputs showed that by applying n-gram and error models, the model proposed in this work obtained the lowest PER and better accuracy for both ISLE and ChildEn. In other words, in our system, the model that produced the best PER is the error language model with a PER of 23\%, which is better than the rate in previous studies.


Previous works in speech recognition report that word error rate (WER) for human annotators on native data is around 5\% in comparison to WER on non-native speech data, which is as high as 30\% for automatic annotations~\cite{Zechner2009}. Our ASR system may not perform better than this human-level performance from human annotators in detecting mispronounced phonemes in non-speech data. The best recent performance report on WER comes from an ETS project, in which the authors trained an ASR system on 800 hours of speech data, and the reported WER was 28.5\%~\cite{Chen2018}.
By considering the error language model, we can see that the ASR system performs better at recognizing the phones. We can see that our ASR system selected the transcription defined in the new adapted lexicon based on our error model. 

Improving ASR will provide more speaking possibilities for learners to learn a new language. Using ASR can improve the traditional class to a more learner-centered environment with less anxiety.  
Apart from the segmental aspect, the suprasegmental aspect also plays a role in pronunciation training, but the ASR for the suprasegmental part is less reliable and still needs more work to improve it.

\section{Conclusions}

In this study, we worked on the language model - at the phone level - to extract the mispronounced phones and trained an acoustic model of native and non-native training examples to detect mispronunciations. We compared the PER of our ASR system using native and non-native speech data to determine our error model hypothesis's validity. Our system's innovation uses error rules in our language model (error model) based on the most common errors seen in L2 learners' speech to improve our detection system. Our error rules were defined based on the common errors of Italian speakers who learn English as their second language.  
The error detection system records the learners’ utterances, and the ASR-based detector will provide the phone-level transcription. As the final step, a list of possible feedbacks can be given to L2 learners based on their pronunciation errors. Our ASR system needs to be adapted to the mother tongue of the L2 learners since different L1s can cause different pronunciation errors.

\bibliographystyle{IEEEtran}

\bibliography{inteerspeech21}


\end{document}